\pdfoutput=1

\documentclass[11pt]{article}

\usepackage[final]{acl}

\usepackage{times}
\usepackage{latexsym}

\usepackage[T1]{fontenc}

\usepackage[utf8]{inputenc}

\usepackage{microtype}
\usepackage{inconsolata}

\usepackage{graphicx}
\usepackage{subfiles}
\usepackage{graphicx}
\usepackage{subcaption}
\usepackage{enumitem}
\usepackage{float}
\restylefloat{table}
\author{
\textbf{Zixin Tang\textsuperscript{1}},~~
\textbf{Janet G. van Hell\textsuperscript{2, 3}}\\
\textsuperscript{1}College of Information Sciences and Technology,\\
\textsuperscript{2}Department of Psychology, \textsuperscript{3}Center for Language Science\\
The Pennsylvania State University\\
\texttt{\{zxtang,jvg3\}@psu.edu}
}

\usepackage{xspace}
\usepackage{comment}
\usepackage{xcolor}
\usepackage{multirow}
\usepackage{booktabs}
\usepackage{multirow}
\usepackage{cleveref}
\usepackage{booktabs}
\usepackage{multirow}

\usepackage{soul}

\title{Learning to Write Rationally:\\How Information Is Distributed in Non-Native Speakers' Essays}

\newcommand{\kenneth}[1]{}
\newcommand{\zixin}[1]{}

\newcommand{\eg}{{\it e.g.}\xspace}
\newcommand{\ie}{{\it i.e.}\xspace}

\begin{document}
\setlength{\abovedisplayskip}{3pt}
\setlength{\belowdisplayskip}{3pt}
\maketitle

\kenneth{Or author response here. We should incorporate all the responses into the final draft: \url{https://docs.google.com/document/d/1rMoH2P06kUyIoMeNjCiqqMvqtDwDQCwK/edit?usp=sharing&ouid=103262082947048789067&rtpof=true&sd=true}}

\kenneth{Editorial note 1: We use ``language production'' to refer to (1) the process of human writing and (2) the written outcome (e.g., the essay's text) interchangeably. Which one is more accurate? We need to pick one and stick with it clearly. I feel quite confused TBH.}

\kenneth{Editorial note 2: This paper uses a lot of phrases like ``language processing'', ``mechanism'', and ``language production'' to describe humans' cognitive process--- In NLP, these phrases, by default, refer to ``computer'' processes. Everytime we refer to human process, we need to specify.}



\begin{abstract}
People tend to distribute information evenly during language production, such as when writing an essay, to improve clarity and communication.
However, this may pose challenges to non-native speakers.
In this study, we compared essays written by second language (L2) learners with various native language (L1) backgrounds to investigate how they distribute information in their non-native L2 written essays.
We used information-based metrics, \ie, word surprisal, word entropy, and uniform information density,
to estimate how writers distribute information throughout the essay to deliver information. 
The surprisal and constancy of entropy metrics showed that as writers' L2 proficiency increases, their essays show more native-like patterns will be in the essay, indicating more native-like mechanisms in delivering informative but less surprising content.
In contrast, the uniformity of information density metric showed fewer differences across L2 speakers, regardless of their L1 background and L2 proficiency, suggesting that distributing information evenly is a more universal mechanism in human language production mechanisms.
This work provides a computational approach to investigate language diversity, variation, and L2 acquisition via human language production. 
\kenneth{You want to add one or a few sentences at the end of the abstract to say what's the positive impact of the work.}
\zixin{not sure what else I can put as positive impacts at this moment haha.}
\end{abstract}


\section{Introduction\label{sec:introduction}}

With the progress of globalization, more people have started acquiring new languages. 
For instance, the proportion of individuals who speak multiple languages daily in the United States has doubled over the past four decades, rising from about one in ten speakers to about one in five~\cite{multilingualpopulation}. 
These rapid changes in linguistic diversity offer unique opportunities but also present challenges for the multilingual population: 
Not all speakers achieve perfect or proficient levels in their non-native languages (L2s) due to various factors, 
including the quantity and quality of exposure to L2s~\cite{leow1998exposureeffects}, 
the length and styles of their acquisition process~\cite{Legault2019L2immersive}, 
and their native language (L1) backgrounds and experiences~\cite{zdorenko2012articlesL2}.
The cognitive mechanisms underlying language use in multilingual speakers may differ from those of native speakers, not only due to variations in proficiency but also because of diverse language backgrounds and experiences ~\cite{1987competingmodel,2005L2competingmodel}.

Many previous studies have explored whether and how speakers with different language backgrounds comprehend and produce languages differently.
For example, Spanish-English speakers may produce ``Spanish-like'' sentences in their English production, where such types of grammar are rarely used or even prohibited in English.  
Most of these studies have reached a similar conclusion: 
for multilingual speakers, representations are integrated across languages, forming a unified system for human language processing~\cite{putnam2018integrated,espengpriming2004}.
Consequently, for individuals who know more than one language, the language(s) that are not seemly involved in the target language production task, can also contribute to and influence comprehension and production processes in the target language, leading to unique patterns in human language processing that can reveal information and knowledge from other languages.

Despite variations in language production among multilingual speakers, the overarching goal of speaking and writing remains the same: to deliver information effectively. 
To achieve this goal, people distribute information evenly across language production, maintaining relatively equal predictability for each upcoming word~\cite{ERC,UID,2021UID}. 
Furthermore, the information carried by a unit of production (\eg, a word)
can be quantified in several ways, including
surprisal~\cite{shannonentropy}, 
entropy~\cite{shannonentropy,ERC}, and 
the uniformity of information distribution (UID)~\cite{UID,2021UID}. 
These metrics help characterize the underlying rules of human language production, which can be summarized as follows:

\begin{itemize}[noitemsep, topsep=0pt]
    \item \textbf{Surprisal Effect:} Processing unexpected information in the produced signal takes longer.
    \item \textbf{Entropy Rate Constancy (ERC):} The rate of information transmitted in a produced unit remains relatively constant across language production.
    \item \textbf{Uniform Information Density (UID):} People prefer to avoid sudden and rapid changes in information density by evenly distributing information across language production.
\end{itemize}

These rules have been substantiated by a wealth of empirical studies.
For instance, 
people need longer time to process unexpected words during comprehension~\cite{ERCRTeffect2013,wilcox2023surprisal}; 
during production, people maintain uniformity of information and constancy of predictability by selecting shorter words~\cite{ERCshorterword2013}, 
repetitive/familiar syntactic structures~\cite{ERCsyntax2016,xu2018information}, or 
faster speech rate~\cite{ERCspeech2017}. Using information-based metrics, prior studies also explored how the complexity of language production changes across language acquisition, and whether we can predict learners’ proficiency based on those changes ~\cite{kharkwal2014surprisal, sanchez2024perplexity, sun2021complexity}.

What remains unknown, despite numerous studies exploring how individuals use these rules to enhance language production, is how L2 speakers 
apply these rules to distribute information in their L2 production---a topic that remains under-researched. 
Given that L2 speakers often exhibit different preferences in lexical selection and syntactic structures compared to native speakers~\cite{espengpriming2004,structuralpriming}
---variations influenced by their language backgrounds---it is reasonable to assume that these differences may result in distinct patterns in their L2 output.
In this paper, we use several well-established metrics from psycholinguistics and information science to investigate how speakers with diverse L1 backgrounds and varying levels of L2 proficiency distribute information in their written production.

\section{Related Work\label{sec:related-work}}

The cognitive mechanisms underlying multilingual language processing represent a significant research topic spanning multiple fields, including
psychology~\cite{kroll2009handbook, schwieter2015handbook},
linguistics~\cite{bhatia2014linghandbook}, and
cognitive neuroscience~\cite{morgan2023neurohandbook, vanhell2023neurocognitive}. As an integrated mechanism covering multiple languages, multilingual speakers demonstrate several typical cognitive and language patterns, such as cross-lingual priming effect~\cite{espengpriming2004, sung2016processing}, cross-lingual cognate effects~\cite{dijkstra2019multilink}, and code-switching effect~\cite{green2014cs}. Other studies explored how multilingualism impacts general cognitive capabilities and neural structures~\cite{baum2014neuroplasticity, birdsong2018plasticity}. \zixin{I try not to use "neuroplasticity" here to avoid confusion. Not sure if NLP people know this specific term.}
Recently, studies also started involving artificial intelligence to explore multilingualism and potential applications for multilingual populations~\cite{zhai2023AIl2learning}, which provides new opportunities to explore and simulate multilingual processes and potential new methods for language education and proficiency assessment.  

While some prior studies take an information-based approach to investigate human language production, few of them specify the nature of their multilingual sample. Some studies offer intriguing evidence regarding cross-lingual production, such as the observation that multilingual speakers switch languages to avoid using uncommon words, demonstrating the surprisal effect~\cite{surprisalcodeswitching}. Specifically, bilingual speakers are more likely to switch languages when the coming words are difficult to predict, leading to a reduction of information density~\cite{myslin2015codeswitch}. Even though some previous works proposed that different mechanisms may exist to help L2 speakers better deliver information in communication~\cite{costa2008L2alignment}, details regarding these mechanisms remain under-researched.

\begin{figure*}[t]
    \centering
    \includegraphics[width=1\columnwidth]{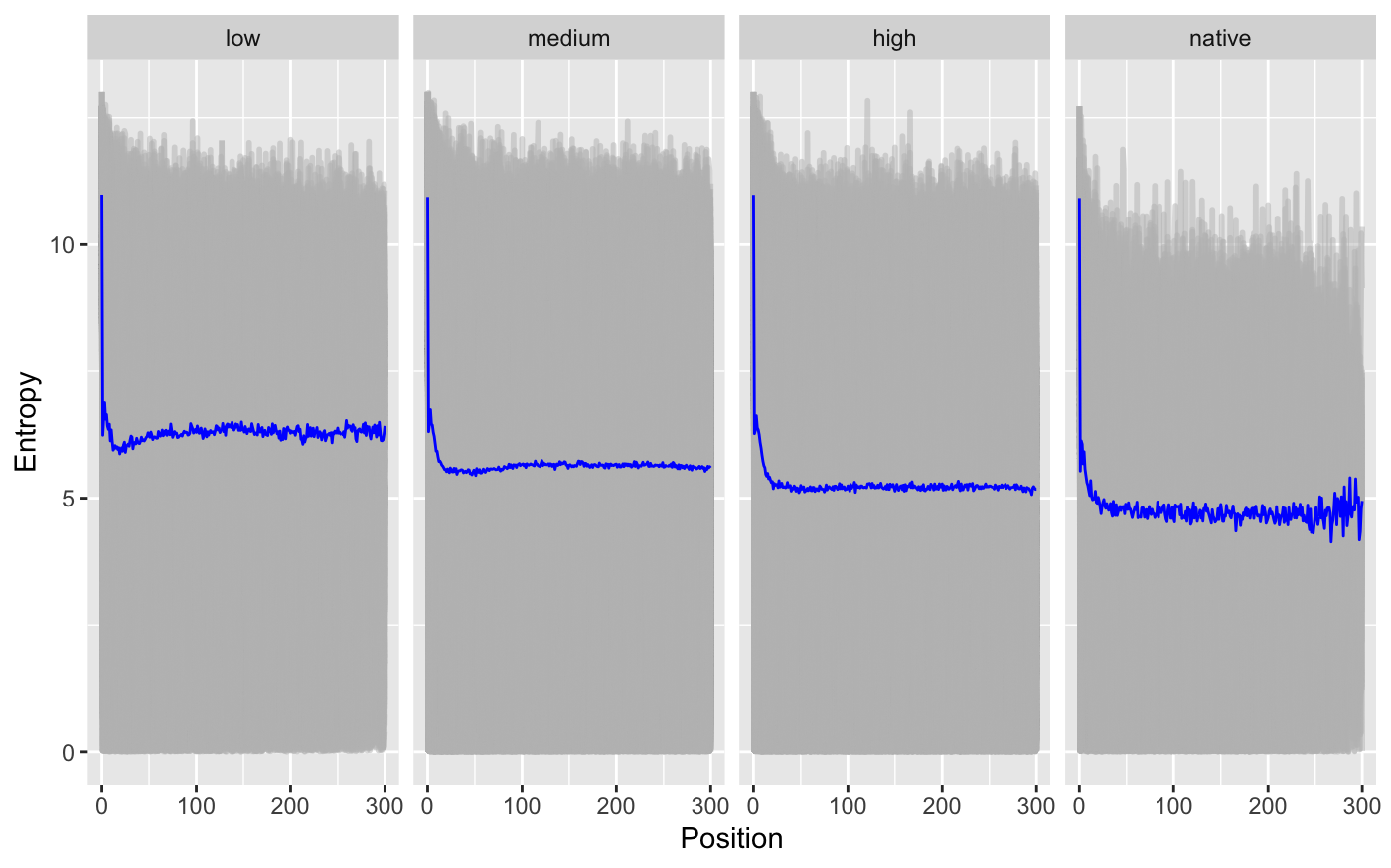}
    \includegraphics[width=1\columnwidth]{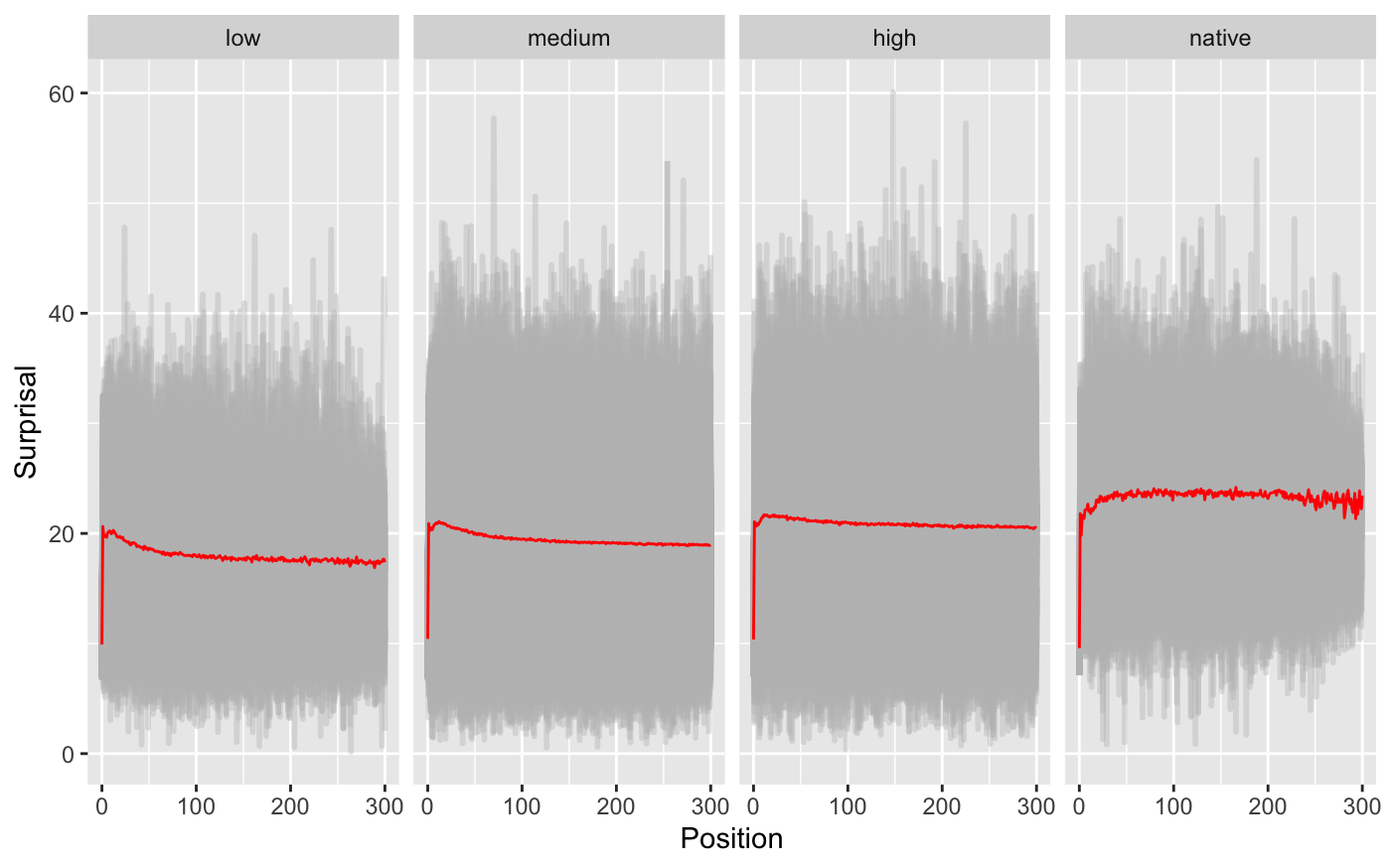}
    \vspace{-3mm}
    \caption{Entropy (left) and surprisal (right) values within written essays, categorized by speaker proficiency. The mean values of both metrics are represented by lines.\kenneth{This figure is too small. Need to make it larger.}}
    \label{fig:position vs. feature}
\end{figure*}

\section{Methods\label{sec:methods}}
\subsection{Materials and Models}
\paragraph{Corpus.}
We used the TOEFL11 corpus~\cite{blanchard2013toefl11} for this study. 
The TOEFL11 corpus contains written essays from actual TOEFL exam takers from 11 different L1 backgrounds. Each L1 category has 1,000 essays, making a total of 11,000 essays in the corpus. 
Speakers are grouped into 3 proficiency groups based on their essay scores.
Detailed information can be found in Appendix~\ref{sec:appendix-corpus descirption}.
Since native English speakers do not typically take the TOEFL exam, we also included 400 essays written by native English speakers from the ICNALE corpus~\cite{ishikawa2013icnale}, which is fewer than any group of L2 learners in the TOEFL11 dataset~\cite{blanchard2013toefl11}. We specifically selected this dataset as a comparison due to its similar setup and data collection process as TOEFL11 corpus: 
the essays in ICNALE corpus are short essays related to discussion-based topics, written within a short time (20-40 minutes). 
Given the similar setup and nature of the written instructions, we used these native speakers' essays to illustrate native-like information distribution patterns.
This inclusion helps in understanding whether and how information distribution varies with changes in speakers' L2 proficiency and L1 backgrounds.

\paragraph{Model.}
Previous corpus-based studies typically analyze the information and language resources within the target corpora. 
However, since the TOEFL11 corpus consists entirely of non-native speakers' written essays, using this method for extracting information measures potentially introduces biases toward non-native-like syntactic structures or lexical selections.
To minimize such biases, we extracted information metrics using pre-trained large language models (LLMs), as these models provide more general and universal estimation regarding tokens' conditional probabilities. 
In this study, we used GPT-2~\cite{GPT2} to tokenize the original essays and convert token-based probability sequences.
We selected GPT-2 as it is an open-access language model without a usage limit. Since GPT-2 is trained based on large-scale web-based materials, it provides a convenient process in capturing general language probability distribution patterns of mainstream language users (English native speakers in our case). 
Because of its openness and transparency, GPT-2 has been used to investigate biases in text generation~\cite{narayanan-venkit-etal-2023-nationality} and information distribution patterns on natural language generation~\cite{gptwho}. Other studies also involved the probability sequences from GPT-2 to predict human behavioral performance~\cite{LLMRT,oh2022entropyRT} and neural behaviors~\cite{michaelov2022GPT2N400,goldstein2022LMsEEG}.


\paragraph{Data Pre-Processing.}
Using the GPT-2 model, we first tokenized the original essays using the GPT-2 tokenizer. -oThe statistic description can be found in Appendix~\ref{sec:appendix-corpus descirption}. \zixin{Should I also include a more specific table as an appendix?}\kenneth{I'm actually thinking we can make a more sophisaced table. BUT let's hold this for now; we can revise the table after sending Janet a draft.}
Each essay had 2 token-based metrics and 3 essay-level metrics to represent the information distribution patterns, detailed extraction processes are introduced in Section~\ref{sec:calculation}. 
Due to the shorter native speakers' essays (250 words, see Table~\ref{sec:appendix-corpus descirption}) and the positively skewed distribution of essay length in the TOEFL11 corpus, the token-based sequences included the first 300 tokens in each essay to balance data sparsity, maintain data completeness, and eliminate less reliable results.



\subsection{Information-Based Metrics\label{sec:calculation}} 
We extracted five metrics from three widely used information-based metrics as follows. 
First, using the token sequences, we obtained the conditional probability \textit{p(w|C)} for each token \textit{w} given all previous context \textit{C}, using GPT-2~\cite{GPT2}.
We then calculated three following three metrics using the probability sequences:

\begin{itemize}[noitemsep, topsep=0pt]
\item 
\textbf{Surprisal:} 
Surprisal~\cite{shannonentropy} measures how much information a signal carries. 
Given the context history (\textit{C}), the surprisal of the \textit{i}-th token is calculated as:

    \begin{equation}\label{surprisal equation}
    S_{i} = -\textit{log}_{2}(p(w_{i}|C_{t<i}))
    \end{equation}
    
In our study, surprisal measures the information density of each token, given the previous context: a lower value indicates a more predictable word. In this study, the surprisal sequence of the first 300 tokens and the mean value of surprisal among all tokens in each essay are extracted as two measures.
    
\item \textbf{Entropy}: 
Entropy measures the expected predictability of the upcoming token~\cite{shannonentropy} through the following equation, given the history of context \textit{C}.
    \begin{equation}\label{entropy equation}
        H_{i} = -\sum_{w\in vocab}(p(w|C_{t<i})\textit{log}(p(w|C_{t<i})
    \end{equation}
Unlike surprisal, 
entropy calculates the average expectancy of the next word before it is produced: a lower value represents a higher certainty regarding the upcoming word. 
In this study, the entropy sequence of the first 300 tokens and the mean value of entropies among all tokens in each essay are extracted.

\item \textbf{UID score:} 
Following previous work on information distribution \cite{UID, 2021UID}, the UID score is measured as the variance of token surprisal, which indicates the information density of each token in the essay. Given the human written production \textit{y}, the UID score represents how uniform the information is distributed across the written production. 

    \begin{equation}\label{UID equation}
        UID(y) = \frac{1}{|y|}\sum_{i}(y_{i} - \overline{y})^{2}
    \end{equation}
    
Based on this equation, a signal with a perfectly even distribution of information receives a 0 UID score.

\end{itemize}

\begin{figure}[t]
    \centering
    \begin{subfigure}[b]{\columnwidth}
         \centering
         \includegraphics[width=1\columnwidth]{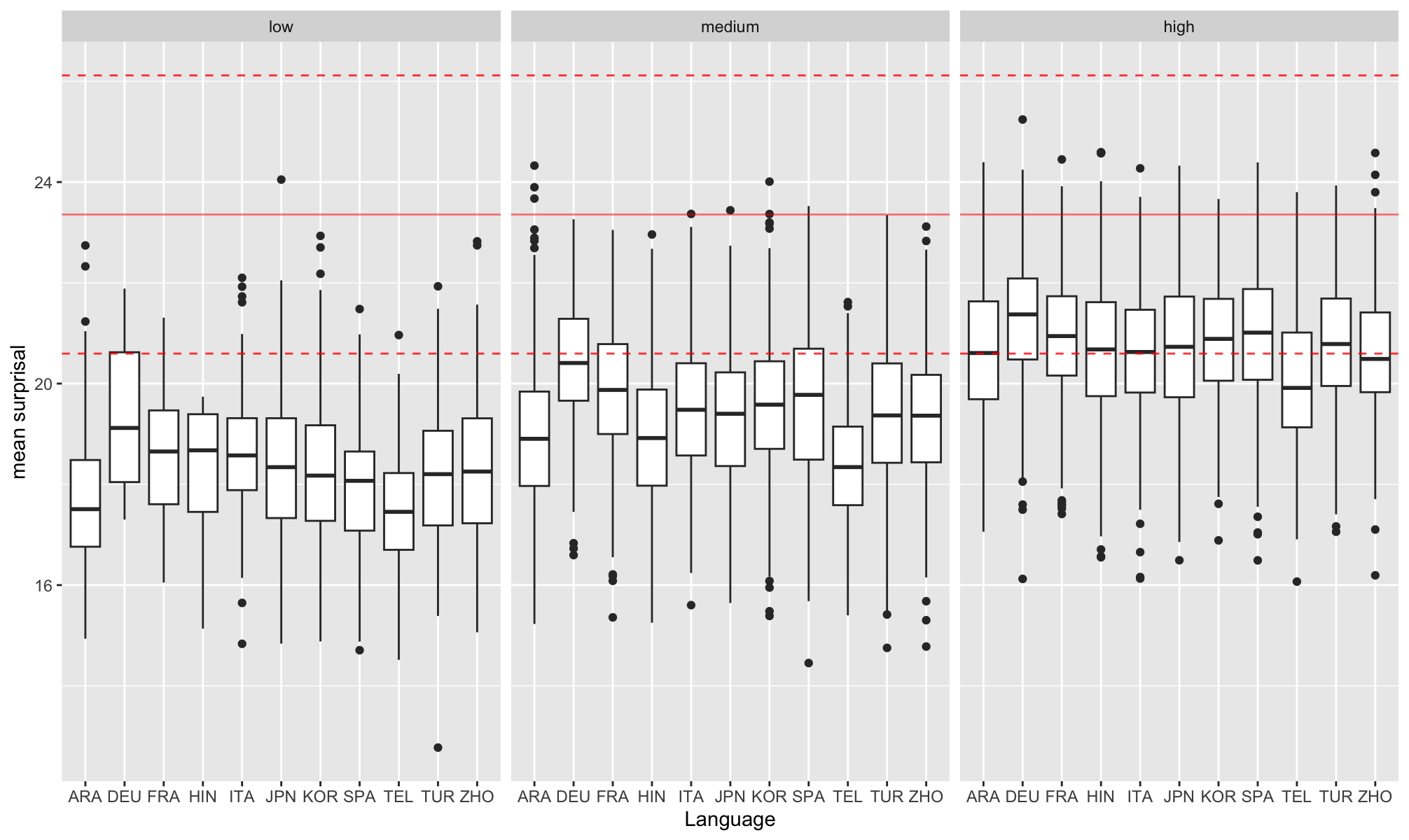}
         \caption{Mean surprisal}
         \label{subfig: mean surprisal}
     \end{subfigure}
    \begin{subfigure}[b]{\columnwidth}
         \centering
         \includegraphics[width=1\columnwidth]{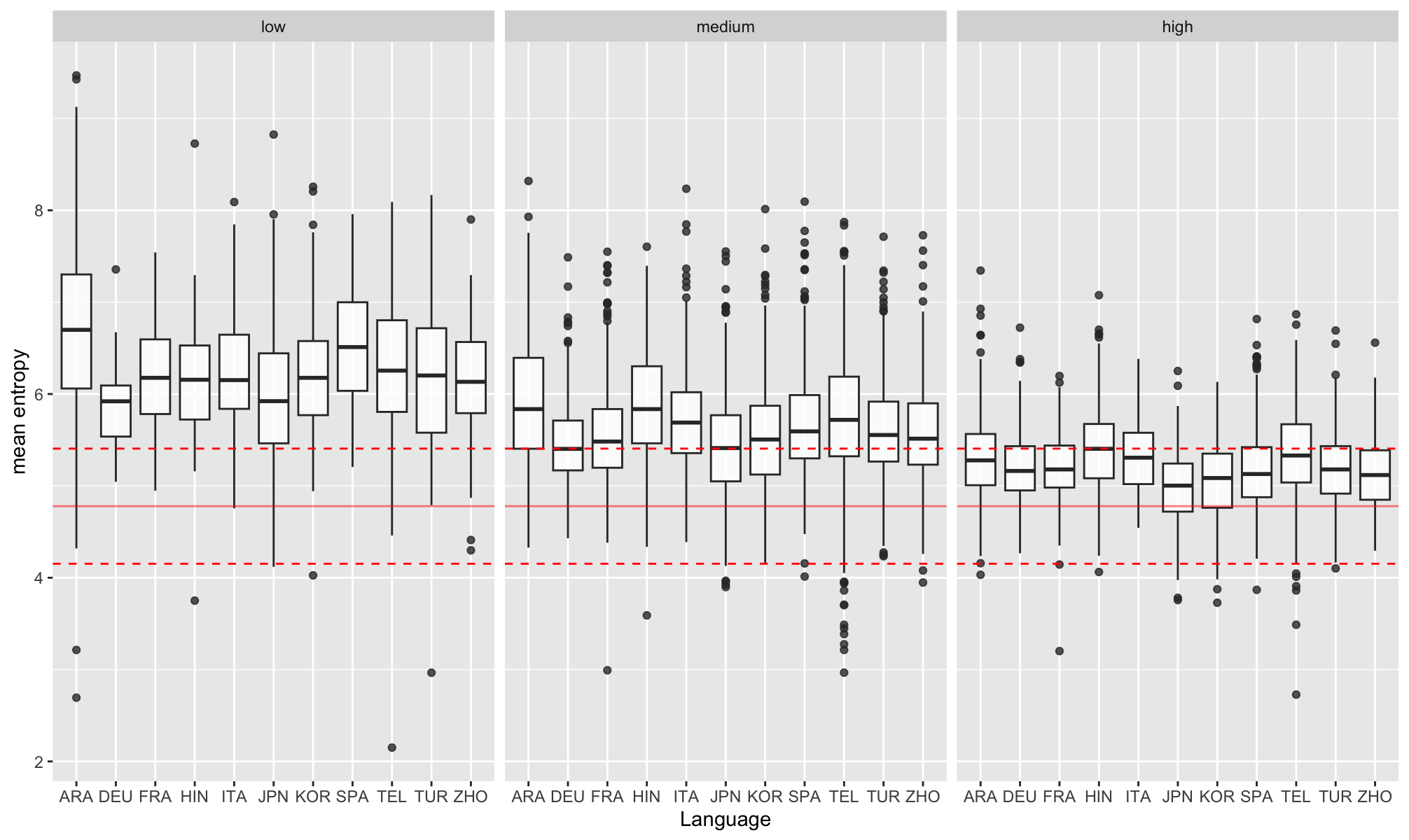}
         \caption{Mean entropy}
         \label{subfig: mean entropy}
     \end{subfigure}
    \begin{subfigure}[b]{\columnwidth}
         \centering
         \includegraphics[width=1\columnwidth]{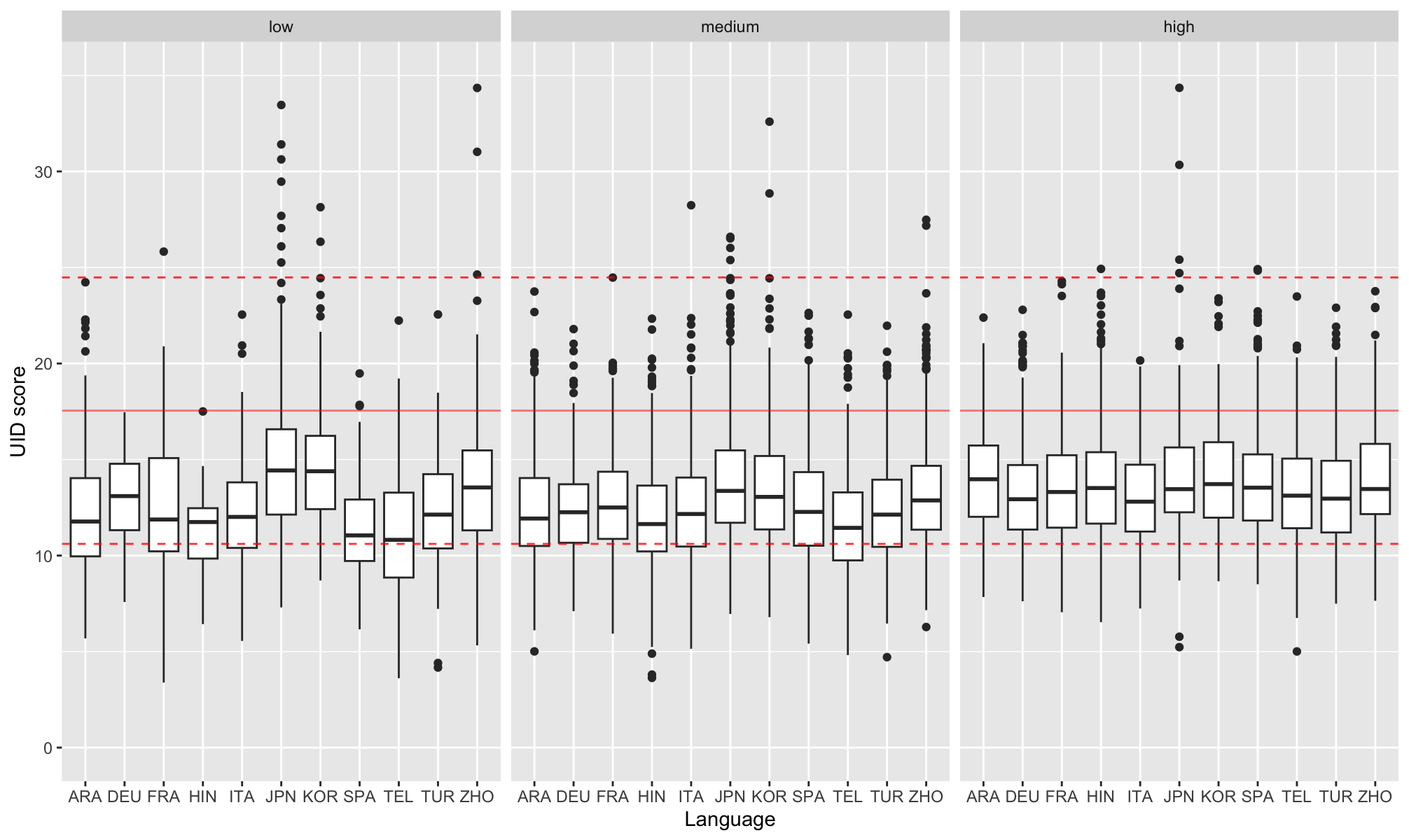}
         \caption{UID score}
         \label{subfig: UID}
     \end{subfigure}
    \vspace{-5mm}
    \caption{Boxplots of information metrics among non-native speakers' essays. Red lines indicate the mean and 95\% distribution among native speakers.\kenneth{This figure is too small. Need to make it larger.}}
    \label{fig:entropy vs. language}
\end{figure}

\section{Results\label{sec:results}}
\subsection{Proficiency vs. Information Distribution\label{sec:result-prof}}
\zixin{S and E have significant predictive power, while UID doesn't due to the poor explained variance in response (R\textsuperscript{2}) and $\eta$}
We fitted two linear mixed-effect models using token-based surprisal and entropy as response variables, token positions and proficiency as fixed effects, and individual essays as random effects. 
We observed a trend towards more native-like patterns, with decreasing entropy values and increasing surprisal values in position-based results as the speaker's proficiency increases (Figure~\ref{fig:position vs. feature} and Table~\ref{table: mixed effect}).
Such a pattern was also observed in the following essay-level analysis (Figure~\ref{fig:entropy vs. language}).
These findings indicate the significance of L2 proficiency in predicting how native-like the information distribution pattern is in L2 production: a higher L2 proficiency is associated with lower uncertainty, but a higher level of informative content. 

Due to the lack of predictive power ($\eta$ = 0.07), there are no significant differences in UID scores regarding speakers with different proficiency levels. Such a pattern can also be observed in Figure~\ref{fig:entropy vs. language}, and will be further discussed in Sec.~\ref{sec:discussion}.
\begin{table}
\centering
\begin{tabular}{lcc}
\hline
\textbf{Proficiency} & \textbf{Surprisal} & \textbf{Entropy} \\
\hline
low      & -3.974\textsuperscript{***}  & 1.256\textsuperscript{***}  \\ 
medium   & -2.739\textsuperscript{***}  & 0.696\textsuperscript{***}  \\
high     & -1.703\textsuperscript{***}  & 0.391\textsuperscript{***}  \\
\hline
\multicolumn{3}{l}{***\textit{p-value} < 0.001}
\end{tabular}
\caption{$\beta$ values of proficiency (native speakers as reference level) of linear mixed effects models.}
\label{table: mixed effect}
\end{table}

\begin{table}
\centering
\begin{tabular}{lccc}
\hline
\textbf{Proficiency} & \textbf{Surprisal} & \textbf{Entropy} & \textbf{UID} \\
\hline
low      & 9.37   & 13.69  & 12.11\\ 
medium   & 70.57   & 34.17  & 21.74\\
high     & 21.74   & 26.64  & 4.89\\
\hline
\end{tabular}
\caption{F-scores regarding each metric in ANOVA analysis with proficiency control.
\zixin{Update: change it to F-score table, move the UID post-hoc to Appendix.}
}
\label{table: F score table}
\end{table}

\subsection{L1 Background vs. Information Distribution}\label{result:L1}
\zixin{I revise the intro part and mentioned L1 background as native language environment and experience.}
Using only L2 speakers' data and essay-based metrics, a one-way analysis of variance (ANOVA) indicated a significant effect of L1 backgrounds on (*** indicates \textit{p} < 0.001): 
\zixin{apa-guideline on reporting ANOVA results}
\begin{itemize}[noitemsep, topsep = 3.5pt]
\item 
\textbf{Mean surprisal}, \textit{F}(10, 10989) $=$ 143.1\textsuperscript{***}, 
\item 
\textbf{Mean entropy}, \textit{F}(10, 10989) $=$ 82.14\textsuperscript{***}, and 
\item 
\textbf{UID}, \textit{F}(10, 10989) $=$ 28.22\textsuperscript{***}. 

\end{itemize}

These effects remained significant when controlling for proficiency (Figure~\ref{fig:entropy vs. language}), indicating that speakers' information distribution patterns are influenced by their L1 background.

Controlling for proficiency, Table~\ref{table: F score table} summarized the variations of essay-level metrics as F-scores in ANOVA analysis.
Medium-proficient L2 speakers show the largest variation in distributing information in terms of all three metrics, while UID showed less variations compared to the other two metrics (Table~\ref{table: F score table}).
This pattern is further discussed in the following sections.


\section{Discussion\label{sec:discussion}}


This study explored how speakers with different L1 backgrounds distribute information in their L2 written production. 
Our results revealed more ``native-like'' trends in metrics such as surprisal and entropy as the speakers' L2 proficiency increased 
In contrast, metrics such as the UID score indicated that L2 writers tend to adhere to the fundamental principles of information distribution, even when they are less proficient in L2. 
These results provide additional insights regarding the learning progress among L2 speakers in language production and communication. 

Language surprisal and entropy emphasize language production from different aspects: Surprisal measures the exact information carried by the incoming word, while entropy estimates the expected certainty about upcoming words.
As demonstrated by native speakers in Figure~\ref{fig:position vs. feature}, speakers want to maximize the information in each word while minimizing the overall expected uncertainty for effective and clearest communication. 
As learners' proficiency in L2 increases, they develop more native-like language production. 
With increased L2 proficiency, they have more L2 resources, which further lead to more advanced, sophisticated, and coherent lexical selection, longer production units, and more complex syntactic structures in their production outcomes~\cite{crossley2020linguisticdevelop, lu2010syntactic,lu2011syntaxcomplexity}. 
Our results provide additional insights through the information distribution among L2 speakers, showing that higher L2 proficiency enables learners to produce language more effectively and efficiently by carrying more information and reducing expected uncertainty in their production.
\kenneth{What prior work or literature could we interact with here?}

Even though we observed significant group differences in mean surprisal and entropy scores among L2 speakers with different L2 proficiency levels and L1 backgrounds, the UID scores showed a different pattern with fewer variations and a more native-like distribution across all proficiency groups (see Figure~\ref{subfig: UID} and Table~\ref{table: F score table}). 
Since UID is associated with the variance of surprisal in language production, the UID variations might indicate that the ability to distribute information evenly might be acquired and generalized as a universal production skill across languages, regardless of how proficient a speaker is in the target language.

\section{Conclusion and Future Work\label{sec:conclusion}}
This paper studies how information is distributed in written essays from native and non-native English speakers using information-based metrics. 
The increasing surprisal and decreasing entropy values showed that proficient L2 speakers distribute information in a more native-like style by maximizing the usage of information channels while reducing the uncertainty of upcoming words. 
In contrast, the UID score showed fewer differences among 
proficiency groups, indicating that maintaining smooth communication channels is a more general skill among human language users.
Future studies can investigate the relationship between linguistic features and information-based metrics regarding speakers' language production, as well as how prior language experiences impact the information distribution patterns. 


\section*{Limitations}

Our study is among the first to explore surprisal, entropy, and uniform information density in L2 English writing in a large group of L2 English speakers with a wide variety of L1 backgrounds and with varying levels of L2 English proficiency.
Here, we outline several limitations of the present work and provide directions for future research. 

Firstly, the dataset contained only basic information regarding speakers’ language background and experience. The only information available in the TOEFL11 dataset is the speakers’ L1. Other crucial details, such as the frequency of L2 usage, duration of L2 acquisition, and the amount of exposure to language(s) other than their L1 and L2 English, are missing. This lack of information restricts the analysis and discussions of underlying causes of the observed variations within each subgroup in the data set, making it challenging to investigate the diversity of language production in depth. We also only explored the information distribution patterns across L2 English learners’ written products, which may restrict the generalizability when dealing with languages from other language families. Future studies may use datasets that include more details regarding language history and the L2 acquisition process, and/or corpora in other languages, to further explore variations in speakers’ language production and information distribution patterns and to better understand the language learning trajectories and language representations in multilingual speakers.

Secondly, our metric calculations may underestimate local changes and fluctuations in information distribution. The essay-level metrics can ignore or underestimate the impact of production length, as longer texts may exhibit larger variations in information density due to the larger number of produced words. In our study, we addressed this issue by analyzing the first 300 tokens in the essays for position-based models. However, this method has a hard cut-off of the essays, potentially leading to incomplete representations of information density distribution. Future studies could address this issue by analyzing shorter production units, such as sentences or paragraphs, to better investigate how information is distributed among L2 learners’ written production.

Thirdly, this study assumes that the probability sequences estimated by LLMs can represent human-like psycholinguistics patterns, which is supported by several studies~\cite{michaelov2022GPT2N400, goldstein2022LMsEEG, michaelov2024GPT3N400}. However, several studies showed that LLMs may not directly represent humans’ mechanisms regarding language comprehension~\cite{mccoy2019LLMheuristics, oh2022LLMweakerpredictor, oh2023LLMpoorerfit}. The differences in “language acquisition” processes between humans and machines can lead to fundamental differences in language representations and mechanisms, even if their final outputs appear similar. Future studies should further investigate the differences in language representations and mechanisms across humans and machines, and examine how such differences can impact the usage of modern computational models in traditional language science research areas.

Lastly, our work focused on computational-based metrics (surprisal, entropy, and UID) and we did not examine more traditional linguistic features, such as specific syntactic constructions. Research has shown that to maintain UID, speakers select specific types of lexical items and syntactic structures when producing languages~\cite{ERCsyntax2016}. In the L2 acquisition process, as proficiency increases, learners have more language resources available to produce language, which leads to more complex, richer, and more appropriate lexical selections and syntactic structures in their language production~\cite{crossley2020linguisticdevelop, lu2011syntaxcomplexity}. Future studies could examine the relationships between computational linguistics metrics and traditional linguistic features for a more complete and detailed understanding of L2 speakers’ acquisition and language production.



\section*{Acknowledgements}
This study is supported by the National Science Foundation (DGE NRT 2125865 and BCS 1734304). 
We thank Ting-Hao `Kenneth' Huang for his generous support and suggestions on revising the manuscript. 
We also thank the anonymous reviewers and the meta reviewer for their helpful comments on the earlier draft. 

\bibliography{custom}

\appendix
\section{Appendix}
\subsection{Corpus Description}\label{sec:appendix-corpus descirption}
\begin{table*}[tbh]
\centering
\begin{tabular}{lccc}
\hline
\textbf{Language} & \textbf{Language family} & \textbf{Portion of essays\textsuperscript{a}} & \textbf{Mean (SD) of essay length\textsuperscript{b}}\\
\hline
Arabic & Afro-Asiatic & 0.274, 0.545, 0.181 & 341.87 (95.21) \\
German (DEU) & Germanic & 0.014, 0.371, 0.615 & 392.06 (73.51)\\
French & Romance & 0.060, 0.526, 0.414 & 372.04 (78.23)\\
Hindi & Indo-Iranian & 0.025, 0.399, 0.576 & 417.42 (86.96)\\
Italian & Romance & 0.145, 0.569, 0.286 & 340.37 (78.90)\\
Japanese& Altaic & 0.207, 0.617, 0.176 & 335.33 (99.16)\\
Korean & Altaic & 0.154, 0.617, 0.229 & 356.48 (97.00)\\
Spanish & Romance & 0.073, 0.502, 0.425& 382.84 (77.35)\\
Telugu & Dravidian & 0.086, 0.595, 0.319 & 418.69 (95.22)\\
Turkish & Altaic & 0.073, 0.561, 0.366 & 373.41 (88.03)\\
Chinese (ZHO) & Sino-Tibetan & 0.090, 0.662, 0.248 & 384.87 (84.44)\\
\hline
\multicolumn{4}{l}{\textit{\textsuperscript{a}of low, medium, and high proficiency speakers.}}\\  \multicolumn{4}{l}{\textit{\textsuperscript{b}mean (SD) of native speakers: 250.72 (30.92).}}
\end{tabular}
\caption{Corpus description.}
\end{table*}
We included the TOEFL11 corpus~\cite{blanchard2013toefl11} and 400 native speakers' essays from the ICNALE corpus~\cite{ishikawa2013icnale} for this study. The detailed information regarding the dataset is listed below, where essay length is measured as GPT-2 tokens.

\subsection{Post-hoc Analysis of Essay-level metrics}
Besides the F-scores from ANOVA analysis, we also conducted the post hoc analysis to investigate the variations of information distribution among L2 English learners with different L1 backgrounds and proficiency. The following tables showed the post hoc analysis results for the surprisal metric (Table~\ref{post-hoc: surprisal}), the entropy metric (Table~\ref{post-hoc: entropy}), and the UID metric (Table~\ref{post-hoc: uid}). Similar to the F-scores result in Section~\ref{result:L1}, we found more significantly different L1 pairs among medium proficiency speakers, indicating these speakers have more variation in terms of information distribution patterns than less and more proficient speakers. Measured as the number of significantly different L1 pairs, the UID metric shows less variation than the surprisal and entropy metrics, suggesting that distributing information evenly when producing written language is a more universal mechanism for human language users. 

\begin{table*}
\centering
\begin{tabular}{lcccccccccc}
\hline
\textbf{Language} & \textbf{ARA}   & \textbf{DEU}    & \textbf{FRA}    & \textbf{HIN}    & \textbf{ITA}    & \textbf{JPN}    & \textbf{KOR}    & \textbf{SPA} & \textbf{TEL}   & \textbf{TUR} \\ \hline
ARA       & -              &                 &                 &                 &                 &                 &                 &              &                &              \\
DEU       & \textbf{1.688} & -               &                 &                 &                 &                 &                 &              &                &              \\
FRA       & \textbf{0.860} & -0.828          & -               &                 &                 &                 &                 &              &                &              \\
HIN       & 0.572          & -1.116          & -0.288          & -               &                 &                 &                 &              &                &              \\
ITA       & \textbf{0.922} & -0.766          & 0.062           & 0.350           & -               &                 &                 &              &                &              \\
JPN       & \textbf{0.685} & -1.003          & -0.175          & 0.113           & -0.237          & -               &                 &              &                &              \\
KOR       & \textbf{0.572} & -1.116          & -0.288          & \textless 0.001 & -0.350          & -0.113          & -               &              &                &              \\
SPA       & 0.247          & \textbf{-1.441} & -0.613          & -0.325          & \textbf{-0.675} & -0.438          & -0.325          & -            &                &              \\
TEL       & -0.176         & \textbf{-1.864} & \textbf{-1.036} & -0.748          & \textbf{-1.098} & \textbf{-0.861} & \textbf{-0.748} & -0.423       & -              &              \\
TUR       & 0.465          & -1.223          & -0.395          & -0.107          & -0.457          & -0.220          & -0.108          & 0.217        & 0.641          & -            \\
ZHO       & \textbf{0.690} & -0.998          & -0.170          & 0.118           & -0.232          & 0.005           & 0.118           & 0.443        & \textbf{0.866} & 0.226        \\ \hline
\multicolumn{11}{c}{(a) Low proficiency}\\
\multicolumn{11}{c}{ }\\

\hline
\textbf{Language} & \textbf{ARA}    & \textbf{DEU}    & \textbf{FRA}    & \textbf{HIN}    & \textbf{ITA}    & \textbf{JPN}    & \textbf{KOR}    & \textbf{SPA}    & \textbf{TEL}   & \textbf{TUR} \\ \hline
ARA       & -               &                 &                 &                 &                 &                 &                 &                 &                &              \\
DEU       & \textbf{1.452}  & -               &                 &                 &                 &                 &                 &                 &                &              \\
FRA       & \textbf{0.902}  & \textbf{-0.550} & -               &                 &                 &                 &                 &                 &                &              \\
HIN       & 0.021           & \textbf{-1.431} & \textbf{-0.881} & -               &                 &                 &                 &                 &                &              \\
ITA       & \textbf{0.526}  & \textbf{-0.926} & \textbf{-0.376} & \textbf{0.505}  & -               &                 &                 &                 &                &              \\
JPN       & \textbf{0.359}  & \textbf{-1.092} & \textbf{-0.543} & \textbf{0.339}  & -0.166          & -               &                 &                 &                &              \\
KOR       & \textbf{0.604}  & \textbf{-0.848} & \textbf{-0.298} & \textbf{0.583}  & 0.078           & 0.244           & -               &                 &                &              \\
SPA       & \textbf{0.662}  & \textbf{-0.790} & -0.240          & \textbf{0.641}  & 0.136           & \textbf{0.302}  & 0.058           & -               &                &              \\
TEL       & \textbf{-0.545} & \textbf{-1.997} & \textbf{-1.447} & \textbf{-0.566} & \textbf{-1.071} & \textbf{-0.904} & \textbf{-1.148} & \textbf{-1.207} & -              &              \\
TUR       & \textbf{0.441}  & \textbf{-1.010} & \textbf{-0.460} & \textbf{0.421}  & -0.084          & 0.082           & -0.162          & -0.220          & \textbf{0.986} & -            \\
ZHO       & \textbf{0.381}  & \textbf{-1.071} & \textbf{-0.521} & \textbf{0.360}  & -0.145          & 0.022           & -0.222          & \textbf{-0.281} & \textbf{0.926} & -0.060       \\ \hline
\multicolumn{11}{c}{(b) medium proficiency}\\
\multicolumn{11}{c}{ }\\
\hline
\textbf{Language} & \textbf{ARA}    & \textbf{DEU}    & \textbf{FRA}    & \textbf{HIN}    & \textbf{ITA}    & \textbf{JPN}    & \textbf{KOR}    & \textbf{SPA}    & \textbf{TEL}   & \textbf{TUR} \\ \hline
ARA       & -               &                 &                 &                 &                 &                 &                 &                 &                &              \\
DEU       & \textbf{0.637}  & -               &                 &                 &                 &                 &                 &                 &                &              \\
FRA       & 0.286           & \textbf{-0.350} & -               &                 &                 &                 &                 &                 &                &              \\
HIN       & -0.012          & \textbf{-0.649} & \textbf{-0.299} & -               &                 &                 &                 &                 &                &              \\
ITA       & -0.035          & \textbf{-0.672} & -0.322          & -0.023          & -               &                 &                 &                 &                &              \\
JPN       & 0.084           & \textbf{-0.553} & -0.202          & 0.097           & 0.119           & -               &                 &                 &                &              \\
KOR       & 0.194           & \textbf{-0.443} & -0.092          & 0.207           & 0.229           & 0.110           & -               &                 &                &              \\
SPA       & 0.314           & \textbf{-0.323} & 0.027           & \textbf{0.326}  & \textbf{0.349}  & 0.229           & 0.120           & -               &                &              \\
TEL       & \textbf{-0.582} & \textbf{-1.219} & \textbf{-0.869} & \textbf{-0.570} & \textbf{-0.547} & \textbf{-0.667} & \textbf{-0.776} & \textbf{-0.896} & -              &              \\
TUR       & 0.155           & \textbf{-0.482} & -0.131          & 0.167           & 0.190           & 0.071           & -0.039          & -0.159          & \textbf{0.737} & -            \\
ZHO       & -0.028          & \textbf{-0.665} & -0.315          & -0.016          & 0.007           & -0.112          & -0.222          & \textbf{-0.342} & \textbf{0.554} & -0.183       \\ \hline
\multicolumn{11}{l}{\textit{\textsuperscript{a} A negative number indicates a smaller mean value for the row L1.}}\\
\multicolumn{11}{l}{\textit{\textsuperscript{b} A bold value indicates a significant difference between row and column L1 (p-value < 0.05).}}\\
\multicolumn{11}{c}{(c) High proficiency}\\

\end{tabular}
\caption{Post-hoc group difference of surprisal metric regarding L1, with proficiency control.}
\label{post-hoc: surprisal}
\end{table*}

\begin{table*}
\centering
\begin{tabular}{lcccccccccc}
\hline
\textbf{Language} & \textbf{ARA}    & \textbf{DEU} & \textbf{FRA} & \textbf{HIN} & \textbf{ITA} & \textbf{JPN}   & \textbf{KOR}   & \textbf{SPA}    & \textbf{TEL} & \textbf{TUR} \\ \hline
ARA & - & &  &  &  &   &   &   &  &  \\
DEU & \textbf{-0.777} & -  &  &  &  &  &  &   &    &   \\
FRA & \textbf{-0.519} & 0.258 & -   &     &    &   &   &  &  &  \\
HIN & \textbf{-0.535} & 0.242 & -0.016 & - &  &  &  &   &  & \\
ITA & \textbf{-0.456} & 0.321 & 0.063  & 0.080 & - &  &  &  &  & \\
JPN & \textbf{-0.718} & 0.059 & -0.199 & -0.183 & -0.262 & -  &  &  &  & \\
KOR & \textbf{-0.523} & 0.254 & -0.004 & 0.013 & -0.067 & 0.196 & - &  &  & \\
SPA & -0.134 & 0.643 & 0.385 & 0.401 & 0.322 & \textbf{0.584} & \textbf{0.388} & - &  & \\
TEL & \textbf{-0.434} & 0.343 & 0.085 & 0.101 & 0.022 & 0.284 & 0.089 & -0.300  & - & \\
TUR & \textbf{-0.543} & 0.234 & -0.023 & -0.007 & -0.087 & 0.176 & -0.020 & \textbf{-0.408} & -0.109 & -  \\
ZHO & \textbf{-0.559} & 0.218 & -0.040 & -0.023   & -0.103  & 0.159 & -0.036      & \textbf{-0.425} & -0.125  & -0.016  \\ \hline
\multicolumn{11}{c}{(a) Low proficiency}\\
\multicolumn{11}{c}{ }\\

\textbf{Language} & \textbf{ARA}    & \textbf{DEU}   & \textbf{FRA}   & \textbf{HIN}    & \textbf{ITA}    & \textbf{JPN}   & \textbf{KOR}   & \textbf{SPA} & \textbf{TEL}    & \textbf{TUR} \\ \hline
ARA               & -               &                &                &                 &                 &                &                &              &                 &              \\
DEU               & \textbf{-0.450} & -              &                &                 &                 &                &                &              &                 &              \\
FRA               & \textbf{-0.342} & 0.108          & -              &                 &                 &                &                &              &                 &              \\
HIN               & -0.035          & \textbf{0.415} & \textbf{0.307} & -               &                 &                &                &              &                 &              \\
ITA               & \textbf{-0.172} & \textbf{0.278} & \textbf{0.170} & \textbf{-0.137} & -               &                &                &              &                 &              \\
JPN               & \textbf{-0.453} & -0.003         & -0.111         & \textbf{-0.418} & \textbf{-0.281} & -              &                &              &                 &              \\
KOR               & \textbf{-0.363} & 0.087          & -0.021         & \textbf{-0.328} & \textbf{-0.191} & 0.090          & -              &              &                 &              \\
SPA               & \textbf{-0.225} & \textbf{0.225} & \textbf{0.117} & \textbf{-0.190} & -0.053          & \textbf{0.228} & \textbf{0.138} & -            &                 &              \\
TEL               & \textbf{-0.184} & \textbf{0.266} & \textbf{0.158} & \textbf{-0.149} & -0.012          & \textbf{0.269} & \textbf{0.179} & 0.041        & -               &              \\
TUR               & \textbf{-0.282} & \textbf{0.168} & 0.060          & \textbf{-0.247} & -0.109          & \textbf{0.171} & 0.081          & -0.057       & -0.098          & -            \\
ZHO               & \textbf{-0.334} & 0.116          & 0.008          & \textbf{-0.299} & \textbf{-0.161} & \textbf{0.119} & 0.029          & -0.109       & \textbf{-0.150} & -0.052       \\ \hline
\multicolumn{11}{c}{(b) Medium proficiency}\\
\multicolumn{11}{c}{ }\\

\hline
\textbf{Language} & \textbf{ARA}    & \textbf{DEU}    & \textbf{FRA}    & \textbf{HIN}    & \textbf{ITA}    & \textbf{JPN}   & \textbf{KOR}   & \textbf{SPA}   & \textbf{TEL}    & \textbf{TUR} \\ \hline
ARA               & -               &                 &                 &                 &                 &                &                &                &                 &              \\
DEU               & \textbf{-0.141} & -               &                 &                 &                 &                &                &                &                 &              \\
FRA               & \textbf{-0.119} & 0.022           & -               &                 &                 &                &                &                &                 &              \\
HIN               & 0.075           & \textbf{0.216}  & \textbf{0.194}  & -               &                 &                &                &                &                 &              \\
ITA               & -0.019          & \textbf{0.122}  & 0.100           & -0.094          & -               &                &                &                &                 &              \\
JPN               & \textbf{-0.347} & \textbf{-0.206} & \textbf{-0.228} & \textbf{-0.422} & \textbf{-0.328} & -              &                &                &                 &              \\
KOR               & \textbf{-0.256} & \textbf{-0.115} & \textbf{-0.137} & \textbf{-0.331} & \textbf{-0.237} & 0.091          & -              &                &                 &              \\
SPA               & -0.143          & -0.002          & -0.024          & \textbf{-0.218} & \textbf{-0.123} & \textbf{0.204} & \textbf{0.113} & -              &                 &              \\
TEL               & 0.014           & \textbf{0.155}  & \textbf{0.133}  & -0.061          & 0.033           & \textbf{0.361} & \textbf{0.270} & \textbf{0.156} & -               &              \\
TUR               & \textbf{-0.135} & 0.007           & -0.015          & \textbf{-0.209} & \textbf{-0.115} & \textbf{0.213} & \textbf{0.122} & 0.008          & \textbf{-0.148} & -            \\
ZHO               & \textbf{-0.184} & -0.043          & -0.065          & \textbf{-0.259} & \textbf{-0.165} & \textbf{0.163} & 0.072          & -0.041         & \textbf{-0.198} & -0.050       \\ \hline
\multicolumn{11}{l}{\textit{\textsuperscript{a} A negative number indicates a smaller mean value for the row L1.}}\\
\multicolumn{11}{l}{\textit{\textsuperscript{b} A bold value indicates a significant difference between row and column L1 (p-value < 0.05).}}\\
\multicolumn{11}{c}{(c) High proficiency}\\

\end{tabular}
\caption{Post-hoc group difference of entropy metric regarding L1, with proficiency control.}
\label{post-hoc: entropy}
\end{table*}

\begin{table*}
\centering
\begin{tabular}{lcccccccccc}
\hline
\textbf{Language}                & \textbf{ARA}     & \textbf{DEU}                  & \textbf{FRA}     & \textbf{HIN} & \textbf{ITA}     & \textbf{JPN}      & \textbf{KOR}       & \textbf{SPA}     & \textbf{TEL}     & \textbf{TUR} \\ \hline
ARA  & -              &                             &                &            &                &                 &                  &                &                &             \\
DEU  & 0.617          & -      &                &            &                &                 &                  &                &                &             \\
FRA  & 0.417          & -0.200                      & -              &            &                &                 &                  &                &                &             \\
HIN  & -0.032         & -0.649                      & -0.449         & -          &                &                 &                  &                &                &             \\
ITA  & -0.054         & -0.671 & -0.471         & -0.023     & -              &                 &                  &                &                &             \\
JPN  & \textbf{2.652} & 2.035                       & \textbf{2.235} & 2.684      & \textbf{2.707} & -               &                  &                &                &             \\
KOR  & \textbf{2.436} & 1.819                       & 2.019          & 2.468      & \textbf{2.491} & -0.216          & -                &                &                &             \\
SPA  & -0.749         & -1.366                      & -1.166         & -0.717     & -0.694         & \textbf{-3.401} & \textbf{-3.185} & -              &                &             \\
TEL  & -1.303         & -1.920                      & -1.720         & -1.271     & -1.248         & \textbf{-3.955} & \textbf{-3.739}  & -0.554         & -              &             \\
TUR & 1.469          & 0.852                       & 1.052          & 1.501      & 1.523          & -1.183          & -0.967           & 2.218          & \textbf{2.772} & -           \\
ZHO & \textbf{2.270} & 1.653                       & 1.853          & 2.302      & \textbf{2.324} & -0.382          & -0.166           & \textbf{3.019} & \textbf{3.573} & 0.801       \\ \hline
\multicolumn{11}{c}{(a) Low proficiency}\\
\multicolumn{11}{c}{}\\

\hline
\textbf{Language}                & \textbf{ARA}     & \textbf{DEU}                  & \textbf{FRA}     & \textbf{HIN} & \textbf{ITA}     & \textbf{JPN}      & \textbf{KOR}       & \textbf{SPA}     & \textbf{TEL}     & \textbf{TUR} \\ \hline
ARA            & \multicolumn{1}{c}{-} &                       &                       &                       &                       &                       &                       &                       &                       &                       \\
DEU            & 0.032                 & \multicolumn{1}{c}{-} &                       &                       &                       &                       &                       &                       &                       &                       \\
FRA            & 0.341                 & 0.308                 & \multicolumn{1}{c}{-} &                       &                       &                       &                       &                       &                       &                       \\
HIN            & -0.232                & -0.265                & -0.573                & \multicolumn{1}{c}{-} &                       &                       &                       &                       &                       &                       \\
ITA            & 0.070                 & 0.037                 & -0.271                & 0.302                 & \multicolumn{1}{c}{-} &                       &                       &                       &                       &                       \\
JPN            & \textbf{1.433}        & \textbf{1.400}        & \textbf{1.093}        & \textbf{1.665}        & \textbf{1.363}        & \multicolumn{1}{c}{-} &                       &                       &                       &                       \\
KOR            & \textbf{1.141}        & \textbf{1.109}        & \textbf{0.801}        & \textbf{1.373}        & \textbf{1.071}        & -0.292                & \multicolumn{1}{c}{-} &                       &                       &                       \\
SPA            & 0.202                 & 0.170                 & -0.138                & 0.435                 & 0.132                 & \textbf{-1.231}       & \textbf{-0.939}       & \multicolumn{1}{c}{-} &                       &                       \\
TEL            & -0.576                & -0.609                & \textbf{-0.917}       & -0.344                & \textbf{-0.646}       & \textbf{-2.009}       & \textbf{-1.717}       & \textbf{-0.779}       & \multicolumn{1}{c}{-} &                       \\
TUR           & -0.046                & -0.079                & -0.387                & 0.186                 & 0.116                 & \textbf{-1.480}       & \textbf{-1.188}       & -0.249                & 0.530                 & \multicolumn{1}{c}{-} \\
ZHO           & \textbf{0.886}                 & \textbf{0.854}        & 0.546                 & \textbf{1.118}        & \textbf{0.816}        & -0.547                & -0.255                & \textbf{0.684}        & \textbf{1.462}        & \textbf{0.933}        \\ \hline
\multicolumn{11}{c}{(b) Medium proficiency}\\
\multicolumn{11}{c}{}\\

\hline
\textbf{Language}                & \textbf{ARA}     & \textbf{DEU}                  & \textbf{FRA}     & \textbf{HIN} & \textbf{ITA}     & \textbf{JPN}      & \textbf{KOR}       & \textbf{SPA}     & \textbf{TEL}     & \textbf{TUR} \\ \hline
ARA            & \multicolumn{1}{c}{-} &                       &                       &                       &                       &                       &                       &                       &                       &                       \\
DEU            & -0.778                & \multicolumn{1}{c}{-} &                       &                       &                       &                       &                       &                       &                       &                       \\
FRA            & -0.450                & 0.328                 & \multicolumn{1}{c}{-} &                       &                       &                       &                       &                       &                       &                       \\
HIN            & -0.246                & 0.532                 & 0.204                 & \multicolumn{1}{c}{-} &                       &                       &                       &                       &                       &                       \\
ITA            & \textbf{-0.905}       & -0.127                & -0.456                & -0.659                & \multicolumn{1}{c}{-} &                       &                       &                       &                       &                       \\
JPN            & 0.182                 & \textbf{0.961}        & 0.632                 & 0.429                 & \textbf{1.088}        & \multicolumn{1}{c}{-} &                       &                       &                       &                       \\
KOR            & 0.036                 & \textbf{0.815}        & 0.486                 & 0.283                 & \textbf{0.942}        & -0.146                & \multicolumn{1}{c}{-} &                       &                       &                       \\
SPA            & -0.038                & \textbf{0.740}        & 0.411                 & 0.208                 & \textbf{0.867}        & -0.221                & -0.075                & \multicolumn{1}{c}{-} &                       &                       \\
TEL            & -0.541                & 0.237                 & -0.091                & -0.295                & 0.364                 & -0.723                & -0.578                & -0.503                & \multicolumn{1}{c}{-} &                       \\
TUR           & -0.672                & 0.106                 & -0.222                & -0.426                & 0.233                 & -0.854                & -0.709                & -0.634                & -0.131                & \multicolumn{1}{c}{-} \\
ZHO           & -0.051                & \textbf{0.727}        & 0.399                 & 0.195                 & \textbf{0.854}        & -0.233                & -0.088                & -0.013                & 0.490                 & 0.621                 \\ \hline
\multicolumn{11}{l}{\textit{\textsuperscript{a} A negative number indicates a smaller mean value for the row L1.}}\\
\multicolumn{11}{l}{\textit{\textsuperscript{b} A bold value indicates a significant difference between row and column L1 (p-value < 0.05).}}\\
\multicolumn{11}{c}{(c) High proficiency}\\

\end{tabular}
\caption{Post-hoc group difference of UID metric regarding L1, with proficiency control.}
\label{post-hoc: uid}
\end{table*}

\end{document}